\documentclass[sigconf,screen]{acmart}
\AtBeginDocument{%
  \providecommand\BibTeX{{%
    \normalfont B\kern-0.5em{\scshape i\kern-0.25em b}\kern-0.8em\TeX}}}

\setcopyright{acmcopyright}
\copyrightyear{2025}
\acmYear{2025}
\setcopyright{acmlicensed}\acmConference[KDD '25]{Proceedings of the 31st ACM
SIGKDD Conference on Knowledge Discovery and Data Mining V.2}{August 3--7,
2025}{Toronto, ON, Canada}
\acmBooktitle{Proceedings of the 31st ACM SIGKDD Conference on Knowledge
Discovery and Data Mining V.2 (KDD '25), August 3--7, 2025, Toronto, ON, Canada}
\acmDOI{10.1145/3711896.3737229}
\acmISBN{979-8-4007-1454-2/2025/08}

\usepackage[utf8]{inputenc} % allow utf-8 input
\usepackage{url}            % simple URL typesetting
\usepackage{booktabs}       % professional-quality tables
\usepackage{amsfonts}       % blackboard math symbols
\usepackage{nicefrac}       % compact symbols for 1/2, etc.
\usepackage{microtype}      % microtypography
\usepackage{float}
\usepackage{natbib}
\usepackage{soul}
\usepackage{graphics}
\usepackage{graphicx}
\usepackage{subfigure}
\usepackage{balance}

\usepackage[ruled,vlined,linesnumbered]{algorithm2e}
\usepackage{paralist}
\usepackage{multirow,multicol,xspace}
\usepackage{amsthm,amsmath}
\usepackage[capitalize,noabbrev,nameinlink]{cleveref}
\usepackage{enumitem}

\renewcommand{\paragraph}[1]{\noindent \textbf{#1}}

\begin{document}

\title{GiGL: Large-Scale Graph Neural Networks at Snapchat}

\author{Tong Zhao*, Yozen Liu*, Matthew Kolodner, Kyle Montemayor, Elham Ghazizadeh$^\dag$, Ankit Batra$^\dag$\\ Zihao Fan, Xiaobin Gao, Xuan Guo, Jiwen Ren, Serim Park, Peicheng Yu, Jun Yu \\ Shubham Vij*, Neil Shah*}
\affiliation{
  \institution{
  Snap Inc., USA
  }
  \country{}
}
\email{*{tong,yliu2,svij,nshah}@snap.com}

\thanks{$*$ Corresponding authors. \\$\dag$ Work done while at Snap Inc.}
% \author{Tong Zhao}
% \affiliation{
%   \institution{
%   Snap Inc. 
%   }
%   \city{Seattle}
%   \state{WA}
%   \country{USA}
% }
% \email{tong@snap.com}

\renewcommand{\shortauthors}{Zhao, et al.}

\begin{abstract}
Recent advances in graph machine learning (ML) with the introduction of Graph Neural Networks (GNNs) have led to a widespread interest in applying these approaches to business applications at scale. GNNs enable differentiable end-to-end (E2E) learning of model parameters given graph structure which enables optimization towards popular node, edge (link) and graph-level tasks.  While the research innovation in new GNN layers and training strategies has been rapid, industrial adoption and utility of GNNs has lagged considerably due to the unique scale challenges that large-scale graph ML problems create. In this work, we share our approach to training, inference, and utilization of GNNs at Snapchat.  To this end, we present GiGL (Gigantic Graph Learning), an open-source library to enable large-scale distributed graph ML to the benefit of researchers, ML engineers, and practitioners. We use GiGL internally at Snapchat to manage the heavy lifting of GNN workflows, including graph data preprocessing from relational DBs, subgraph sampling, distributed training, inference, and orchestration.  GiGL is designed to interface cleanly with open-source GNN modeling libraries prominent in academia like PyTorch Geometric (PyG), while handling scaling and productionization challenges that make it easier for internal practitioners to focus on modeling. GiGL is used in multiple production settings, and has powered over 35 launches across multiple business domains in the last 2 years in the contexts of friend recommendation, content recommendation and advertising.  This work details high-level design and tools the library provides,  scaling properties, case studies in diverse business settings with large-scale graphs up to \emph{hundreds of millions} of nodes, \emph{tens of billions} of edges, and \emph{hundreds} of node and edge features, and several key lessons learned in employing graph ML at scale on large social data.  GiGL is open-sourced at \url{https://github.com/Snapchat/GiGL}.
\end{abstract}


\maketitle

\section{Introduction}
\label{sec:intro}

Graphs are ubiquitous, representing a wide variety of real-world data across domains such as social networks and recommendation systems. In recent years, Graph Neural Networks (GNNs) have emerged as powerful tools for learning from graph data \cite{kipf2016semi, veličković2018graph, ying2018graph}. 

Nevertheless, when deploying GNNs in industrial contexts, such as large-scale user modeling and social recommendation systems, scalability poses a significant challenge. Most commonly used libraries in the research community assume that graph topology, node and edge features, can fit easily within the CPU (or even GPU) memory of a single machine. However, just storing a graph with 100 billion edges as an example (not uncommon in practice \cite{ching2015one}) requires 800 GB of memory assuming 32-bit integer IDs, even before accounting for node and edge features or the need to represent graphs with more nodes than 32-bit limits.  Such limitations make it difficult to scale GNN models in real-world, billion-scale graphs, which we routinely handle at Snapchat.

To address these challenges, we developed GiGL (Gigantic Graph Learning), an internal library that our internal researchers and practitioners use to efficiently train and infer GNNs on large-scale graphs with tens to hundreds of billions of edges. GiGL seamlessly integrates with popular GNN libraries like PyTorch Geometric (PyG)~\citep{Fey2019pyg}, allowing users to leverage  existing modeling codes without introducing new syntax for defining GNN layers. Specifically designed for large-scale graph learning tasks, GiGL offers:
\begin{itemize}[leftmargin=*,topsep=0pt]
    \item Support for both supervised and unsupervised applications, including node classification, link prediction, and self-supervised representation learning. 
    \item Abstracted interfaces for integration with popular machine learning frameworks such as PyTorch and TensorFlow.
    \item Compatibility with widely-used graph ML libraries like PyG for flexible GNN modeling.
    \item Utilities for efficient data transformation, pipeline management, and orchestration in large-scale deployments and recurrent applications.
\end{itemize}
At its core, GiGL abstracts the complexity of distributed processing for massive graphs, enabling users to focus on model development while the library handles data transformation, subgraph sampling, and persistence. This empowers users to iterate on graph models at scale with limited concern to infrastructure and scale constraints.

We built GiGL to address our internal needs over the years. GiGL currently powers all GNN-based graph machine learning workflows at Snapchat, supporting critical product features such as friend recommendations, lens recommendations, spam and abuse detection, and content recommendations. Across multiple use-cases, we use GiGL to generate either task-specific embeddings or general-purpose user embeddings, which serve as inputs for embedding-based dense retrieval (EBR) systems and/or as added dense-features in lower-funnel ranking models. 
Yet, we believe there are opportunities for GiGL to serve a broader community. GiGL addresses a critical need for machine learning researchers, engineers, and practitioners: enabling large-scale exploration and experimentation with state-of-the-art GNN models on industry-scale graphs, while compatibly interfacing with familiar open-source libraries which enable fast model iteration and development like PyG.  While these libraries are more commonly used in contexts with few millions of nodes and edges, they benefit from strong community support and cutting-edge GNN research implementations. GiGL bridges this gap by providing tooling around handling distributed graph scaling challenges, utilities, orchestration and more to handle  heavy lifting, allowing researchers to continue leveraging the modeling tools they know and trust. Hence, this work accompanies an open-source release of GiGL\footnote{\url{https://github.com/Snapchat/GiGL}}. We will evolve this offering to support the broader graph ML community, and are happy to accept contributions.

In this work, we first discuss GiGL's development, components and capabilities, and scaling considerations. Readers will understand design rationale, technical tools used, and infrastructure considerations in using the library.  We next discuss multiple ways in which GiGL-trained GNN embeddings have facilitated solving large-scale ML problems at Snapchat, spanning critical tasks for social platforms like friend recommendation, content recommendation and ads ranking; across these domains, GiGL has powered over 35 launches which span all elements of GNN modeling, spanning graph definition, modeling levers, loss function design. We discuss key choices in each context, experimental results, and significant topline metrics movements across the business.  Finally, we share lessons learned from working with large-scale GNNs across these applications, and our perspective and understanding on GNNs at scale as the  community has evolved over the years. These may be instructive to academic researchers who are interested in peeking into industrial challenges, as well as practitioners wanting to advance GNN modeling in their own contexts.

\vspace{-0.05in}
\section{Preliminaries and Related Work}
\label{sec:preliminary}

\paragraph{Graph Neural Networks.}
GNNs are at the current forefront of graph machine learning research. 
They are designed to transform input nodes into compact vector representations, which can be used for node, link and graph-level tasks. GNNs commonly use the layer-wise message-passing design~\citep{kipf2016semi,veličković2018graph,xu2018powerful,Gao2018LargeScaleLG,wu2020comprehensive,ma2021unified}, in which nodes progressively aggregate and process information from immediate neighbors, enabling convolution over the graph with multiple layers. With the message-passing design, the embedding of each node can be computed given its $k$-hop subgraph for a $k$-layer GNN. The process of generating the $k$-hop subgraphs is often referred as subgraph sampling.

\paragraph{GNN Scalability.}
Most GNN applications in the research landscape occur at a single-machine scale, owing to the size of benchmark datasets~\cite{hu2020open}.  In industrial contexts, graphs may surpass such limits in terms of number of nodes, edges and features, which complicates GNN training and inference due to data dependency in the message-passing design and the need for careful minibatching~\citep{duan2022comprehensive,xue2024large}. 
Recent literature discusses multiple directions to train GNNs with large-scale graphs, including sampling methods~\citep{hamilton2017inductive,chen2018fastgcn}, staleness-based methods~\citep{fey2021gnnautoscale,xue2024haste},  distillation~\citep{zhang2021graph,guo2023linkless}, quantization~\citep{ding2021vq}, condensation and coarsening~\citep{jin2021graph, tsitsulin2023graph}, pre-training~\citep{han2022mlpinit, borisyuk2024lignn}, distributed training~\citep{md2021distgnn, lin2023comprehensive}, and many more~\citep{xue2024large}.
These techniques have been commonly adopted as directions in GNN scaling research, and are applied to training in larger benchmark graphs (e.g., {\small\texttt{ogbn-products}}~\citep{hu2020open} with 2.4M nodes and 61M edges). Nonetheless, in large-scale applications, some of these techniques meet various challenges owing to inefficient implementations, misalignment between academic and industrial assumptions, and diverse graph properties \cite{palowitch2022graphworld} -- for example, graphs we handle at Snapchat are routinely have 10-50B edges. 

\paragraph{Industrial Applications.}
Adopting GNNs has been a key focus in multiple industrial contexts: Pinterest discusses using GNNs to learn pin-board embeddings in an early work~\cite{ying2018graph} which adopts GraphSAGE with hard-negative sampling.  Google showcases the use of GNNs for traffic time forecasting~\cite{derrow2021eta}.  LinkedIn applies GNNs for feed and job recommendations~\cite{borisyuk2024lignn, liu2024linksage}. Uber applies GNNs for food and restaurant discovery~\cite{jain2019food}. At Snapchat, we explored the use of GNNs in friend recommendation~\cite{shi2023embedding, kung2024improving, sankar2021graph}, story ranking~\cite{tang2022friend} and engagement forecasting~\cite{tang2020knowing}. 

Solutions at scale broadly fall into two groups: \emph{real-time} subgraph sampling, and \emph{tabularization}: The former requires maintaining a shared graph state, like an in-memory engine or graph database, which is queried during training and inference.  The latter pre-computes all subgraph samples to cloud storage, which are used for subsequent training and inference via an extract-transform-load (ETL) job; this removes need for shared graph state, akin to standard data-parallel setup common in other ML workflow.
GraphLearn~\citep{zhu2019aligraph}, ByteGNN~\cite{zheng2022bytegnn}, DeepGNN~\cite{samylkin2022deepgnn} and GraphStorm~\cite{zheng2024graphstorm} adopt realtime sampling via various graph backends and partitioning strategies, following ideas by \citet{chiang2019cluster, karypis1997metis}. Conversely, AGL~\cite{zhang13agl}, TF-GNN~\cite{ferludin2022tf}, and MultiSAGE~\cite{yang2020multisage} adopt tabularization.  Both strategies have pros and cons, which we will discuss further in \cref{sec:lesson-infra}.

The library we discuss in this work, GiGL, adopts the tabularization approach by default, but is not constrained to it, and has a configurable real-time backend which allows interfacing with tabularization backends via Spark and graph databases, as well an in-memory graph engine. We have used GiGL successfully internally at Snapchat with different backends, and this flexibility is empowering for challenges at different technical scale, freshness and complexity. We hope our work adds to the literature in this space via \emph{(i)} offering the community a diverse technical perspective, \emph{(ii)} sharing unique lessons learned and business impact earned over years of working on this problem domain, and \emph{(iii)} case studies on billion-scale tasks at Snapchat.  We emphasize that our goal in open-sourcing GiGL and sharing details in this work is to empower the community with helpful tooling and guidance, rather than advocate for GiGL as a superior alternative to other approaches.

\vspace{-0.05in}
\section{Design and Interface}
\label{sec:framework}
\begin{figure}[tbp]
    \centering
    \includegraphics[width=0.45\textwidth]{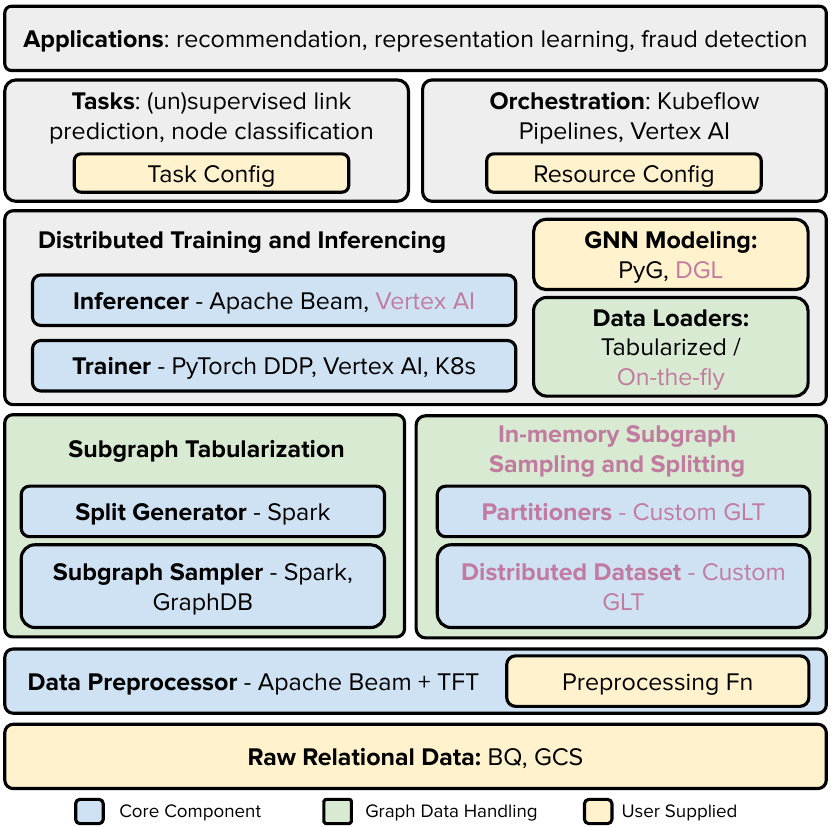}
    \vspace{-0.1in}
    \caption{GiGL framework components; {\color{purple}magenta} items are work-in-progress.}
    \label{fig:gigl}
    \vspace{-0.1in}
\end{figure}

\begin{figure*}[t]
    \centering
    \includegraphics[width=0.95\textwidth]{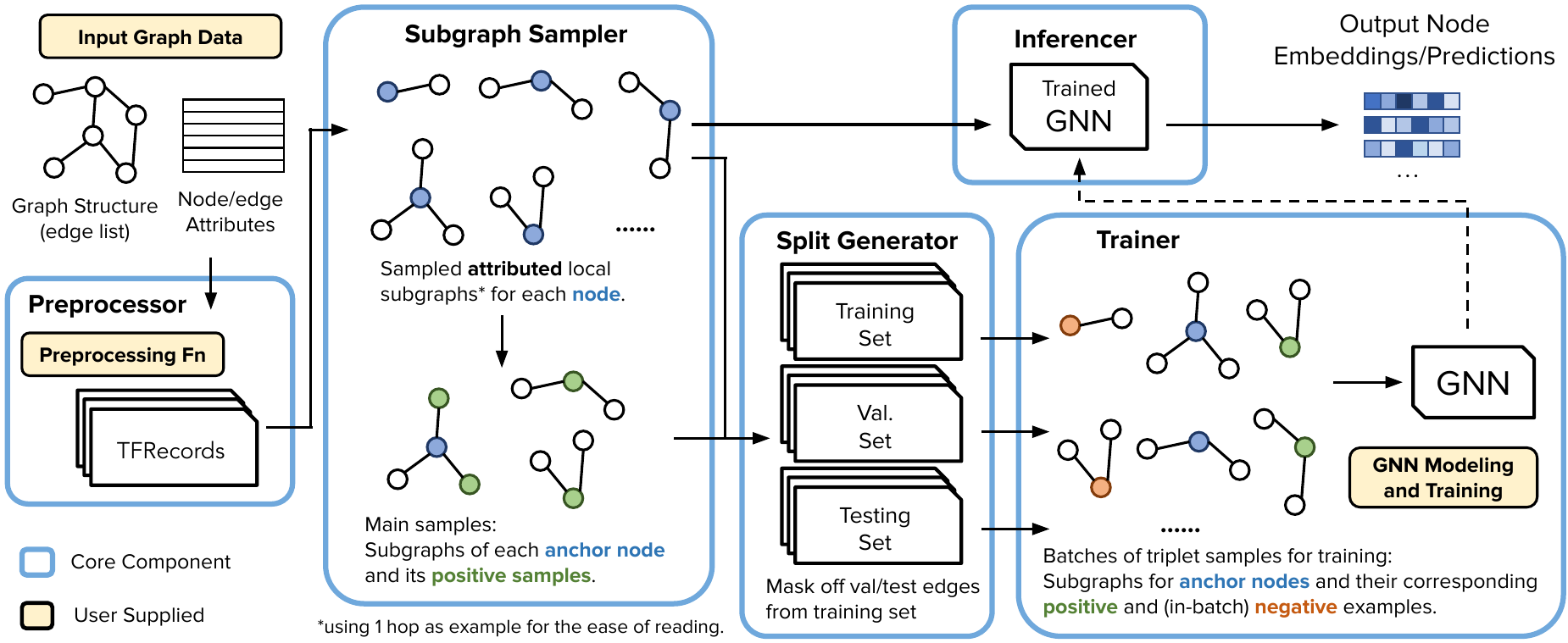}
    \vspace{-0.1in}
    \caption{A example GiGL workflow with tabularized subgraph sampling for the task of link prediction, in which the model is trained with triplet-style contrastive loss on a set of anchor nodes along with their positives and (in-batch) negatives. The intermediate results of each component are stored on a cloud storage service such as GCS.}
    \label{fig:framework}
    \vspace{-0.05in}
\end{figure*}

GiGL is designed with horizontal scaling across many compute resources in mind, making it well-suited for deployment on both custom compute clusters and cloud environments. While vertically-scaled GNN solutions (hosting graph and training on one very large machine) are viable to an extent, their scalability is inherently limited\footnote{We discuss our experience moving away from a vertically-scaled approach in \cref{sec:lesson-infra}.}, and is a bottleneck in environments where data volume can fluctuate across tasks, and grow rapidly with user growth. 

\cref{fig:gigl} provides an overview of the GiGL architecture. As discussed earlier and illustrated in the graph data handling components of \cref{fig:gigl}, GiGL is designed to support two strategies for managing industry-scale graph data: tabularization, and real-time subgraph sampling. Since both workflows share the same orchestration framework and most core components, we begin by introducing GiGL through the tabularization workflow in \cref{sec:components}. We then discuss an in-memory implementation of the real-time subgraph sampling workflow in \cref{sec:glt}, followed by an overview of the shared user interface used across both in \cref{sec:interface}.

\subsection{Components and Orchestration}
\label{sec:components}
We begin by introducing the tabularization scheme, which has been a core part of GiGL and has been in use in multiple domains inside Snapchat with significant success over the past two years. \cref{fig:framework} provides a visual overview of GiGL’s workflow for a sample link prediction task, trained with a triplet-style contrastive loss.
We next outline the design of GiGL pipelines, which are comprised of five main components and orchestration capabilities\footnote{Detailed documentation of each component can be found at \url{https://snapchat.github.io/GiGL/docs/user_guide/overview/architecture.html}}:

\paragraph{Data Preprocessor} (DP)
is a distributed feature transformation pipeline backed by TensorFlow Transform (TFT)\footnote{\url{https://www.tensorflow.org/tfx/transform}},  leveraging the Beam\footnote{\url{https://beam.apache.org}} runner.  TFT is widely used in industry for large-scale data preprocessing, encompassing feature scaling, normalization, handling of categorical features, and more.  DP reads and transforms input graph topology (edge-lists) and raw node and edge features from relational data sources (in our case, BigQuery or Google Cloud Storage) into the \texttt{TFRecord} format.  Users are able to define distributed transforms like one-hot-encoding, normalization, scaling,   imputation and filtering over the graph.  We provide bindings to run DP on Google Cloud Dataflow\footnote{\url{https://cloud.google.com/products/dataflow}} (with AWS support planned).

\paragraph{Subgraph Sampler} (SGS)
is a distributed batch workflow which enables custom graph sampling logic to facilitate tasks like node classification and link prediction. At its core, SGS generates $k$-hop subgraphs for each node in an input graph, with added sampling flexibility.  Since training GNNs requires message-passing subgraph topology and feature data for \emph{all} nodes involved in the loss calculation, SGS supports workflows which also require added positive/negative sampling to generate complete training examples.  For node-level tasks, each sample contains the $k$-hop subgraph and node label. For edge-level tasks like link prediction, SGS generates samples with anchor, positive and negative nodes' subgraphs. Notably, label nodes can also be custom or user-defined, as shown in the second block of \cref{fig:framework}.

GiGL has two separate tabularization backends \emph{Pure-ETL} and Graph Database (\emph{GraphDB}) backed. The Pure-ETL backend strictly takes advantage of constructs in batch-processing pipelines like {\small\texttt{Join}} and {\small\texttt{GroupByKey}}. We implemented this in Apache Spark~\cite{salloum2016big} on Scala. The method leverages repeated joins on edge-list data, with a hydration step to enrich node and edge features.  Notably, our Spark implementation is a re-write of a previous Beam on Dataflow version, which resulted in huge (${\geq}80\%$) cost and runtime improvements.  The GraphDB backend uses an Apache Spark on Scala job which queries a graph DB instance post-ingestion of DP output. Internally, we leveraged {\small\texttt{NebulaGraph}} \cite{wu2022nebula}.  Currently, we utilize the GraphDB backend for internal heterogeneous GNN workflows to support most-flexible neighbor sampling (via imperative graph query language) and compatibility with internal graph infrastructure, but our abstracted design enables easy vendor-swapping.  The Pure-ETL backend is mainly used for simpler homogeneous GNN workflows with limited sampling complexity.  Careful partitioning, caching and resource-management ensure this job's reliability. Aligning support across the backends is work-in-progress.

\paragraph{Split Generator} (SG)
is a distributed data splitting routine to generate globally consistent train/validation/test splits according to flexible split strategies (transductive, inductive, and custom user-defined).  SG also uses Spark on Scala, which was a re-write from Beam on Dataflow for cost and runtime reasons (similar to SGS). 

\paragraph{Trainer}
is a distributed model training job that consumes data from SG for model training, validation, and test loss computation. % to generate model artifacts. 
For each training batch, workers fetch multiple SG output samples and collate them into a single batch subgraph, which is 
fed into user-defined training logic written in familiar modeling code (e.g., PyG). Collation is abstracted-away, such that users can operate directly with simple full-batch training loops.  Moreover, collation can be modified through a translation layer to support other modeling frameworks (e.g., DGL, TF-GNN) -- this is planned work. Trainer outputs a model artifact which can be used in Inferencer, along with relevant evaluation metrics. 

\paragraph{Inferencer}
is a distributed inferencing component which generates embeddings and/or class predictions dictated by user code. Unlike Trainer, Inference uses Beam on Dataflow for parallel CPU/GPU inference. Interestingly, we found that CPU inference offers easier scaling with GNN models given their light parameter footprint. 

\paragraph{Orchestration.}
A common practitioner workflow is the configuration of an end-to-end pipeline which runs components in sequence (see \cref{fig:framework}) to power batch inference.  
GiGL leverages workflow management software to schedule the component jobs in order. GiGL currently provides tooling to launch pipelines on Kubeflow\footnote{\url{https://www.kubeflow.org}} and VertexAI\footnote{\url{https://cloud.google.com/vertex-ai}}. For easier orchestration, we introduce two additional mini-components which run before pipeline-start: \emph{(i)} a Validation Checker which verifies input configuration correctness and path-existence to avoid wasting development and compute cycles, and \emph{(ii)} a Config Populator, which takes template configurations, and populates it with frozen asset paths (transformed data, splits, model artifact, inferences) to simplify and centralize asset tracking. Path persistence prior to pipeline execution enables retrying individual components idempotently, fault tolerance, and amortization of individual component outputs for other runs: e.g., multiple Trainers using the same SG output during hyperparameter tuning.

\subsection{Real-time Subgraph Sampling}
\label{sec:glt}
While tabularization provides a scalable and effective approach for subgraph generation, it has some limits (discussed in \cref{sec:lesson-infra}) when greater flexibility and adaptive use of a graph is required during training.  Real-time subgraph sampling workflows enable this flexibility for speed, at the expense of amortization potential.  To enable such workflows, we integrate and build from a custom version of {\small\texttt{GraphLearn-for-PyTorch}}\footnote{\url{https://github.com/alibaba/graphlearn-for-pytorch}} (GLT).  While sharing the same orchestration, data reading and transformation (DP) logic, our customized GLT backend partitions the graph into a distributed in-memory dataset across machines.  The graphs are semi-randomly partitioned: nodes are shuffled across machines, and adjacent edges are collocated based on (customizable) adjacent source or destination node.  Hence, all nodes are able to access all 1-hop neighbors within-machine, and the associated node and edge features are also collocated.  Once this dataset is established, training and inference workflows are able to generate subgraphs via gRPC calls across machines for multi-hop sampling.  During training, we adopt a similar strategy as in SG to enable globally consistent distributed splits to mask out relevant data, and can flexibly use CPU or GPU training.  This logic operates within the logical scope of the Trainer and Inferencer components, and similar to the tabularization scheme: trainer outputs a model artifact and inference outputs final embeddings or predictions to a sink (BigQuery, or Google Cloud Storage, internally). We have successfully used this scheme internally, and open-source support is currently work-in-progress.

\begin{table*}[!htbp]
\centering
\caption{Component-level runtime comparison of GiGL on different graphs that vary at scale. Note that only internal-less-feat and internal-full have the same graph structure, making the MRR numbers incomparable across all other combinations.}
\label{tab:scale}
\vspace{-0.1in}
\resizebox{0.9\textwidth}{!}{
\begin{tabular}{l|cccccc|c} 
\toprule
 & \multicolumn{6}{c|}{Runtime (mins)} & Offline \\
Dataset & Total & Data Preprocessor & Subgraph Sampler & Split Generator & Trainer & Inferencer & MRR \\ 
\midrule
MAG240M & 8h 43m & 1h 28m & 1h 41m & 37m & 4h 24m & 30m & 0.747 \\
\cmidrule(lr){1-1} \cmidrule(lr){2-7} \cmidrule(lr){8-8}
Internal-small & 5h 19m & 36m & 32m & 9m & 3h 44m & 18m & 0.602 \\ 
Internal-less-feat & 9h & 2h 10m & 1h 50m & 28m & 3h 25m & 1h 7m & 0.704 \\
Internal-less-edge & 8h 41m & 1h 34m & 2h 7m & 39m & 3h 9m & 1h 12m & 0.807 \\
Internal-full & 11h 24m & 2h 31m & 2h 29m & 50m & 4h 4m & 1h 30m & 0.820\\
\bottomrule
\end{tabular}
}
\vspace{-0.1in}
\end{table*}

\subsection{User Interface}
\label{sec:interface}
Apart from the raw relational data, users are asked to specify two configuration files which dictate the GiGL workflow which runs: a \emph{Resource Config} and a \emph{Task Config}; see \cref{appx:config} for links to examples of them, respectively, and our official user guide\footnote{\url{https://snapchat.github.io/GiGL/\#configuration}} for further details and guidelines.

\paragraph{Resource Config.} This configuration specifies the underlying compute resource details which GiGL components should aim to provision, including number of machines, machine spec, disk, memory, CPU and GPU requests.

\paragraph{Task Config.} GiGL is designed with modeling flexibility in mind, enabling users to customize key components through a dependency injection-like paradigm.  Users can provide or override core logic by implementing custom interfaces across various components:

\begin{itemize}[leftmargin=*]
    \item \textbf{Data Preprocessor.} Users can define a custom class with data transformation logic using TFT preprocessing functions for features, or use common defaults within GiGL.
    \item \textbf{Split Generator.} Users can define custom data splitting strategies or use common defaults within GiGL.
    \item \textbf{Trainer and Inferencer.} Users can easily define custom model definitions as in PyG or use common defaults within GiGL.
\end{itemize}

We provide working examples of component logic to guide users while maintaining flexibility for extensive customization.

\section{Experimentation and Scaling}
\label{sec:evaluation}

In this section, we present offline evaluations of GiGL on both public benchmarks as well as production graphs in different sizes to showcase time and resource efficiency of GiGL. We experiment with an internal homogeneous graph that is used for friend recommendation, which has $\sim${900M} nodes, $\sim${16.8B} edges, 249 node features, and 19 edge features. Due to space limit, we leave the comprehensive dataset details in \cref{appx:dataset}.

\cref{tab:scale} showcases the component-level runtime comparison of GiGL running on the internal graph and one public benchmark, MAG240M\footnote{We also offer example codes for running MAG240M end-to-end with GiGL at: \url{https://github.com/Snapchat/GiGL/tree/main/examples/MAG240M}}~\citep{hu2021ogblsc}. We can observe that GiGL is able to finish end-to-end on the full industry-scale graph within 12 hours, making it feasible for product use-cases. Additionally, from the time comparison of the different versions, we can observe that the number of edges generally has a bigger impact on the runtime, aligning with GNNs' time complexity which is linear with the number of edges~\citep{wu2020comprehensive}. Beyond these graph scaling experiments, we also conducted additional offline experiments of GiGL with varying resources and setups, which are included in \cref{tab:scale-het,tab:resource} in \cref{appx:experiments}.

\section{Business Impact and Case Studies}
\label{sec:internal}

GiGL is widely used inside Snapchat to power multiple graph learning usecases. Along with the business applications, we also discuss how we utilize the key advantages of GiGL, such as the flexibility, scalability, and reusability. Additionally, we showcase a couple key features of GiGL that were inspired by our production needs.

\subsection{Friend Recommendation}
\label{sec:friending}

Snapchat's friend recommendation system is a major internal user of GiGL.  Snapchat powers a large social network with hundreds of millions of daily active users, striving to help users discover their real-life friends on the platform. A classic multi-stage retrieval and ranking system is employed to power this large-scale recommendation system which entails recommending potential friends for each user from the entire user base~\citep{shi2023embedding}. Prior to GiGL, the retrieval system primarily exploited principles of graph locality, using discrete graph traversal algorithms to retrieve friend candidates (e.g. friends-of-friends) which serves as candidates in the heavy ranker.

One early usecase of GNNs at Snap was the embedding of this social graph.  In particular, we constructed a friendship graph defined over recently active users, where nodes represent users and edges indicate friendships, with node features indicating  user profile properties, and trained an unsupervised GraphSAGE model with margin loss~\citep{hamilton2017inductive} to generate embeddings for all users on a daily cadence. These embeddings enabled the deployment of a robust Embedding-based Retrieval (EBR) system~\citep{shi2023embedding} powered by efficient Approximated Nearest Neighbor (ANN)~\citep{shi2023embedding} search, which has since become the largest funnel for our friend retrieval stage. 
Additionally, aligned with \citet{sankar2021graph}, we found that incorporating such GNN embeddings as additional features in the ranker further improved online metrics, demonstrating that topology information is beneficial even for precision-focused models. % (further discussed in \cref{sec:lesson-model}).

\begin{table}[t]
\centering
\caption{Several important GiGL launches in Snapchat friend recommendation and their launch impacts highlights on (global or regional) core business metrics such as new friends made, suggestion to new friends made with bi-directional communication, daily pipeline cost savings, etc.}
\label{tab:friending}
\vspace{-0.1in}
\resizebox{0.47\textwidth}{!}{
\begin{tabular}{l|c} 
\toprule
\textbf{Launches} & online improv.  \\ 
\midrule
GraphSAGE on friend graph (initial launch of EBR) & +10\%, +11.6\%, +7\% \\ 
Graph definition update: engagement graph & +8.9\%, +5.8\%, +10.8\% \\
Graph definition update: graph augmentation & +1.0\%, +2.1\%  \\
Model update: GAT & +6.5\%, +2.92\% \\
Loss update: retrieval loss & +6.2\%, +8.8\% \\
Loss update: multi-task training  & +1.2\%, +1.3\%  \\
Task update: supervised LP with user-defined labels & +2.2\%, +1.8\%  \\
EBR scheme update: Stochastic EBR & 10.2\%, 13.9\% \\
\bottomrule
\end{tabular}
}
\vspace{-0.1in}
\end{table}

The initial launch of graph embeddings using GiGL was a major success for Snapchat's friend recommendation system, leading to continuous investments into this joint effort. In the past two years, GiGL has supported more than 20+ successful product launches in multiple areas of the friend recommendation stack. \cref{tab:friending} highlights several of these launches, showcasing either notable findings on GNN usage in friend recommendation or important feature developments in GiGL with application-based learnings:

\paragraph{Graph Definition Update.} 
Real-world data is often noisy, and graph ML in  production systems rarely comes with a ``ready-to-use'' graph. Instead, we rely on leveraging abundant log data containing various types of information.
% \footnote{All user data at Snap is processed in compliance with privacy regulations and laws, such as the EU's General Data Protection Regulation (GDPR).}. 
Initially, we used the friendship graph, which seemed to align well with the friend recommendation use case. 
However, we soon realized that social circles evolve over time, and recent interactions are more indicative of friend recommendations. To address this gap, we introduced a simple graph sparsification strategy of updating the graph to an engagement graph, with nodes still represent users, but edges are defined only between users who recently interacted. This update not only significantly improved performance but also reduced costs, as the engagement graph is far much sparser. In following updates, we also investigated data augmentation techniques~\citep{zhao2021data,zhao2022graph} to enhance the low-degree users, and our recent research NodeDup~\citep{guo2024node} is also on an internal testing roadmap.

\paragraph{Model Update.} 
As discussed in \cref{sec:framework}, GiGL offers key advantages in modeling flexibility by leveraging existing model implementations from external libraries. This enabled us to experiment with all graph convolution layers available in PyG at the time, and we identified attention-based models (e.g. GAT~\citep{velivckovic2017graph}) as the best-performing for our task. 
Over time, similar experiments were conducted periodically and led to multiple model updates, including transitions to GATv2 layers~\citep{brody2021attentive} and the addition of DCN layers~\citep{wang2017deep}. In particular, we hypothesize such attention-based models are especially useful for two main reasons: \emph{(i)} they can improve model expressivity via added parameterization, \emph{(ii)} they can well-handle scale-free graphs with skewed degree distributions (common in social networks), where not all neighbors have equivalent significance in modeling (e.g. defining convolutions based on strong-engagement user relationships can provide stronger signal vs. compared to randomly selected user relationships).

\paragraph{Loss Update.} 
Similar to modeling flexibility, GiGL also offers custom loss-function definition, with several ready-to-use implementations.  For the friend recommendation EBR usecase, we found the Sampled Softmax-based Retrieval Loss~\citep{jean2014using,yi2019sampling,wu2024effectiveness} to be suitable in application, and achieved outsized business gains given its simplicity. Additionally, inspired by recent research~\citep{ju2022multi}, we further adopted a multi-task learning approach which adopted ideas in self-supervised graph representation learning~\cite{kolodner2024robust}: we combined retrieval loss with feature-reconstruction and whitening-decorrelation losses, based on findings in ParetoGNN~\citep{ju2022multi} that these could yield stronger general-purpose representations. Both auxiliary losses resulted in successful product launches.

\paragraph{Task Update.} 
To further enhance GNN embeddings' quality for the task of friend recommendation, we use GiGL's ``supervised'' link prediction setup (not commonly observed in academic literature), 
and replaced the sampling of training edges with user-defined positive (and optionally negative) supervision edges. For example, we used newly established friendships made over next-$k$ days as positive labels while keeping the engagement graph as the message-passing graph. This approach improved business metrics, improved representation quality compared to a purely self-supervised setup, and opened new possibilities for incorporating hard negative signals, such as user ignores and blocks into the training process.

\paragraph{EBR Scheme Update.} A key finding from our GNN embedding iterations for friend recommendation EBR is that ANN seed selection significantly impacts performance. 
We naturally started with each user's own embedding as the query for retrieving potential friend candidates. However, we observed that this approach could limit candidate diversity and miss relevant connections.
To address this,  we developed stochastic EBR~\citep{kung2024improving}, which retrieves candidates using embeddings of a user’s friends instead of their own. 
This method effectively broadened the search space and captured richer social signals.
This approach delivered substantial performance gains in business metrics and has since become the default scheme for GNN-based friend recommendation EBR at Snapchat.

\begin{table}[t]
\centering
\caption{GiGL launches in Snapchat content recommendation and their launch impacts highlights on core business metrics such as view time, total time spent on the platform, subscription sum, etc.}
\label{tab:content-online}
\vspace{-0.1in}
\resizebox{0.46\textwidth}{!}{
\begin{tabular}{l|c} 
\toprule
\textbf{Launches} & online improv.  \\ 
\midrule
Spotlight EBR with GNN embeddings & +0.55\%, +1.54\%, +2.02\% \\ 
Discover EBR with GNN embeddings & +0.35\%, +1.02\%, +3.17\% \\
\bottomrule
\end{tabular}
}
\vspace{-0.1in}
\end{table}

\begin{table}[t]
\centering
\caption{Heterogeneous GNN embedding offline iterations on content recommendation.}
\label{tab:content-offline}
\vspace{-0.05in}
\resizebox{0.47\textwidth}{!}{
\begin{tabular}{l|cc} 
\toprule
\textbf{Setup} & MRR & HR@1  \\ 
\midrule
Base model: 2 layer GAT, sample 15 neighbors per layer & 0.39 & 0.24 \\ 
+ feature normalization & 0.54 & 0.40 \\
+ further feature eng. & 0.59 & 0.46 \\
+ increase to 3 layers, double training data- & 0.64 & 0.53 \\
+ \# of neighbor sampling to 20 & 0.68 & 0.57 \\
+ \# of neighbor sampling to 30, half hidden dim. & 0.70 & 0.59 \\
% + edge def. update to comb. of multiple pos. interaction & 0.80 & 0.74 \\
\bottomrule
\end{tabular}
}
\vspace{-0.05in}
\end{table}

\subsection{Content Recommendation}

Beyond friend recommendation, content recommendation is another key application with abundant relational data that benefits from GNN modeling. The interactions between users and short videos (Spotlight\footnote{\url{https://www.snapchat.com/spotlight}} and Discover\footnote{\url{https://www.snapchat.com/discover}} tab) naturally form bipartite graphs, with user nodes connected to video nodes through various interactions. By incorporating additional metadata, such as content creators, this structure extends to a heterogeneous graph, effectively representing relationship across multiple types of entities. 

\paragraph{Graph Definition Update.}
Before we implemented the GraphDB backend for sampling heterogeneous graphs, we first constructed a video-to-video co-engagement graph to leverage GNN embeddings for content recommendation.
User-video interactions are even more long-tailed in our setting than user-user ones, with many videos receiving millions of positive engagement signals. To address this imbalance, we used the Jaccard index with a threshold (over the sets of co-engaging users) to sparsify the graph, resulting in a more balanced structure which facilitates GNN learning.  
These embeddings were used in a video-to-video EBR system to retrieve similar videos that users had engaged with positively. This application was successfully launched in both of our short video products with measurable business metrics improvements (as shown in \cref{tab:content-online}).

\paragraph{Heterogeneous Support.}
With the heterogeneous graphs support by GraphDB-based subgraph sampling, we then pursued a modeling task designed for a similar use-case, but with a user-engage-video bipartite graph, which allowed us to also leverage user node features and engagement features on the edges.
The embeddings generated by the heterogeneous graph are undergoing both offline and online studies for both video-to-video retrieval and user-to-video retrieval use cases. \cref{tab:content-offline} exemplifies a few observations during our offline iterations, highlighting the improvements brought by feature engineering and increasing number of GNN layers in product usages.  

\begin{table}[t]
\centering
\caption{Homogeneous GNN embedding offline iterations on product ads recommendation.}
\label{tab:ads-offline}
\vspace{-0.1in}
\resizebox{0.47\textwidth}{!}{
\begin{tabular}{l c c c|ccc} 
\toprule
\textbf{Setup} & \# edges &  neighbors size & MRR & Precision & Recall  \\ 
\midrule
Control (graph embed.) & 60B & -& -  & 0.1523 & 0.3220 \\ 
2 layer GAT (v1) & 3.36B &  15 & 0.871 & 0.1138 & 0.3239 \\
2 layer GAT (v2) & 6.46B  & 15 & 0.959 & 0.1505 &  0.3819 \\
2 layer GAT (v3) & 6.46B & 40 & 0.980 & 0.1523 & 0.4109	 \\
\bottomrule
\end{tabular}
}
\vspace{-0.1in}
\end{table}

\subsection{Ads Recommendation}
User-ad engagement is often sparse, noisy, and influenced by various factors. For instance, in a mobile setting, users may unintentionally click on ads when their actual intention was to skip them. Such complexity is further compounded by the presence of multiple in-app ad inventories (e.g., Story Ads, Discover Ads), diverse advertiser objectives (e.g., swipes, clicks, installs, purchases), and different ad products (e.g., websites, apps, physical products). The primary goal of ads modeling is to maximize the utilization of available signals across these varying products, objectives, and inventories. Given the inherently heterogeneous nature of ads engagement, we explore testing of GNN modeling with GiGL on both homogeneous and heterogeneous graphs in different ads product verticals.

\paragraph{Graph Definition Update.}
For product ads recommendation, similar to content recommendation, we constructed a homogeneous graph with product IDs as nodes, and various types of co-interactions as edges. We also adopted a thresholded Jaccard index to achieve a more balanced structure for effective GNN learning.  Additionally, we incorporated contextual embeddings, such as product text and image embeddings, as well as product metadata (price, currency, brands), enabling the model to capture both co-engagement similarity and product-type similarities. The resulting product embeddings are then utilized in EBR systems for product retrieval.  \cref{tab:ads-offline} compares results with the current productionized model. GNN v1 utilizes product text, image embedding and product metadata as features, and GNN v2-v3 utilizes text embedding with product metadata. The precision and recall are computed for 1\% of the randomly sampled fixed set of users' future conversion events (add-to-cart, purchase, etc.) We can observe that with 10\% of the training data, a homogeneous GNN model achieves precision parity with the control model, but with higher recall. As with other applications, we observed that the performance increases as the number of training edges and neighbor fan-out increases. 

For web ads recommendation, we constructed a heterogeneous graph with both web ads and users as nodes, connected by edges representing different interaction types.
Node features include user demographics,  user past engagements, and web ads metadata (e.g., domains). To mitigate the impact of power nodes, where a few dominant entities disproportionately influence conversions, we pruned edges and capped node degrees to 10K in the graph. The pruning helps maintain a balanced graph structure, which is easier for GNNs to learn from.  In particular, our results consistently support that careful graph data definition is an important knob to tune in training. Both use-cases showed strong offline performances and are under-going further online studies for internal testing.

\paragraph{Learning from Partially Observable Graph.}
As users on Snapchat can opt-out from personalized ads recommendation, the interaction data between users and ads are only partially available for model training and inferencing. Hence, we model multiple partial graphs between users and different types of ads (e.g., product ads, app ads, web ads) into a joined heterogeneous graph with multi-task learning rather than multiple individual partial graphs, allowing us to have more control and learn from cross-type ads engagement signals. Furthermore, we also explored graph data augmentation~\citep{zhao2021data} and structure learning~\citep{jin2020graph} techniques on such partial graphs.

\begin{table}[t]
\centering
\caption{Other areas benefiting from social information provided by GNN embeddings learned by GiGL, with their launch impacts on product-specific business metrics.}
\label{tab:other-online}
\vspace{-0.1in}
\resizebox{0.47\textwidth}{!}{
\begin{tabular}{l|c} 
\toprule
\textbf{Applications} & online improv.  \\ 
\midrule
% Drug-related content detection & +10.9\%, +6.9\%, +36\% \\ 
Community guideline violation detection & +10.9\%, +6.9\%, +36\% \\ 
Bad actor detection & +25.9\%, +27.1\% \\
EBR for Display Name Search (DNS) & +6.23\%, +4.65\% \\
Social Graph Embeddings  in DNS Heavy Ranker & +1.01\%, +0.98\%, +0.75\% \\
Social Graph Embeddings in Lens Ranking & +0.54\%, +0.51\%, +0.62\% \\
\bottomrule
\end{tabular}
}
\vspace{-0.1in}
\end{table}

\subsection{Broader Representation Learning Applications}
\label{sec:broader}

Through our experience training GNNs, we discovered that the relational information encoded by GNNs is often valuable not only for a target domain but also for related domains within Snapchat. For example, incorporating social-graph user embeddings from our friend recommendation system as additional features into multiple ML tasks within the business (e.g. bad actor detection), we achieved notable performance gains, likely due to the effective social-prior encoded in the embeddings. \cref{tab:other-online} summarizes several such product launches where existing embeddings were successfully reused.

Looking ahead, the multiple entities and relations across different domains on a social network can be captured in one single large-scale heterogeneous graph.
Powered by our recent research~\citep{ju2022multi}, we aim to develop cross-domain multi-task graph learning models that learn embeddings beneficial to many products simultaneously. While such a pipeline can be computationally expensive, it remains cost-effective as the resource costs are shared across multiple product teams, maximizing both efficiency and business impact.

\section{Lessons and Opportunities}
\label{sec:roadmap}
In this section, we discuss lessons and opportunities informed by our experience with large-scale graph ML. We expect this guidance to help academic researchers as well as practitioners building out their own graph ML functions.

\subsection{Lessons: Infrastructure}
\label{sec:lesson-infra}

\paragraph{Vertical vs. Horizontal Scaling.}  Our early explorations in GNN modeling at industry-scale occurred when modern GNN libraries were limited, and ``spatial GNNs'' were still emerging~\citep{hamilton2017inductive, ying2018graph}. To handle large graphs, we employed a vertical-scaling solution, leveraging the highest-memory cloud machines available at the time (GCP's \texttt{n1-highmem-96}, with 96 vCPU and 624GB RAM).  We stored graph topology in CSR format, node features as a large tensor, and used Ray \footnote{\url{https://github.com/ray-project/ray}} for both shared-memory and multiprocessing, enabling a producer-consumer setup across CPU workers and GPU trainers.

While this prototype powered our early applications, it soon faced challenges due to multiple scaling factors: 
\emph{(i)} Snapchat's rapid user growth, increasing graph size, 
\emph{(ii)} the pursuit of richer node and edge features, and 
\emph{(iii)} the need of adopting more advanced newly-proposed GNN models. 
The reliance on vertical-scaling handicapped our ability to deal with these factors, forcing compromises in model, feature and graph scale. This led us to invest in horizontally scalable architectures as the next-gen GNN modeling solutions at Snapchat (discussed in \cref{sec:framework}). Today, although the ceiling of vertical-scaling has increased with larger machines, we opt for horizontal designs which offer greater elasticity at different scales.

\paragraph{GNN Libraries vs. Hand-rolled Convolutions.} In our early efforts, libraries like PyG \cite{Fey2019pyg} and DGL \cite{wang2019dgl} were nascent, and not well-suited for large-scale industrial use. Hence, we developed graph sampling and convolution operations in native PyTorch, using models like GraphSAGE \cite{hamilton2017inductive} and (neighbor-sampled) GCN \cite{kipf2016semi}.
This approach required managing complex multi-dimensional tensors\footnote{E.g., tensor with shape of \texttt{(batch\_size, n\_nodes, n\_nodes, n\_node\_feat)} for 2-layer convolution.} and intricate reshaping for message passing.
While effective initially, these custom solutions were challenging to adapt to advanced GNN models. Hence, when re-architecting, we take advantage of advances in frameworks like PyG and DGL which abstract away low-level convolution details. GiGL now benefits from the continuous improvements contributed to these libraries, closing the gap between cutting-edge research and production applications.

\paragraph{Tabularization vs. Real-time Graph Sampling.} 
GiGL is designed to support both \emph{tabularized} and \emph{real-time} subgraph sampling workflows. Tabularization enables:
\emph{(i)} cost amortization for parameter tuning and multi-user workflows,
\emph{(ii)} easy integration with traditional recommender data sources to enable co-trained GNNs, and
\emph{(iii)} decoupling graph scaling from GNN modeling through translation layers.
However, it introduces challenges such as \emph{(i)} storage footprint due to data duplication, \emph{(ii)} need for careful resource management of ETL jobs, and \emph{(iii)} careful balancing of I/O and GPU utilization. In contrast, real-time sampling allows:
\emph{(i)} adaptive graph usage during training (e.g., \cite{han2022mlpinit, yoon2021performance}), and
\emph{(ii)} flexibility in swapping graph engine or database backend.
Its limitations include the need to \emph{(i)} repeat sampling for each forward iteration and \emph{(ii)} need for careful concurrency and availability management of backend~\cite{borisyuk2024lignn}. While many internal applications rely on tabularization, careful backend optimization can yield significant cost savings for these large-scale workflows on a per-run basis. These trade-offs may shift as ETL processes and amortization strategies evolve.

\subsection{Lessons: Modeling}
\label{sec:lesson-model}

\paragraph{Offline vs. Online Performance.} 
As with many ML applications, offline-online discrepancies are common. Hence, we often employ post-processing logic to also monitor task-relevant (non-differentiable) surrogate metrics which can help prune unpromising offline candidates early.

\paragraph{Graph Data Sensitivity.} While most GNN research focuses on improved modeling, we find that graph data definition is underlooked and can outweigh model iteration in impact.  
In real-world applications, graphs can often be refined in various ways, as exemplified in \cref{sec:friending}.
In particular, while past work emphasizes graph densification \cite{zhao2021data, zhao2022graph, han2022g} to improve downstream GNN performance, 
we found that signal quality is the key for better embeddings, resulting with \emph{sparser} graph in most cases.
Favorably, we can achieve better performance at reduced compute cost. Moreover, embedding quality for low-degree nodes can be improved with \emph{targeted} augmentation with signal quality guardrails \cite{wang2023topological, guo2024node}: notably, falling back to weaker-signal edges when strong-signal edges are too-sparse.

\paragraph{Shallow Graph Embeddings (SGE) vs. GNNs.} Modern graph ML research mainly focuses on GNNs, and is predated by advancements in shallow graph embedding (SGE) methods like DeepWalk \cite{perozzi2014deepwalk} or LINE \cite{tang2015line}.  While academic evaluation often paints a picture in which GNN methods clearly outperform SGE methods, we observe in multiple internal applications that they are \emph{mutually} useful.  \citet{dongseesaw} points out this complementarity; in particular, SGEs can outperform GNNs in cases where we have limited node or edge-features (avoiding dimensional collapse), or heterophilic signals, where feature-similarity is less predictive.

% \paragraph{Multi-Stage Ranking.} We find GNN embeddings to be especially powerful sources in retrieval phase of multi-stage ranking funnels across applications.  Yet, GNN embeddings can \emph{also} provide additive value in downstream rankers. \cite{sankar2021graph} observes similar phenomena: in particular, it is valuable for optimization goals for the two sets of embeddings to differ.

\vspace{-0.05in}
\subsection{Opportunities}

\paragraph{New Advances in Link Prediction.} Standard message-passing neural networks (MPNNs) are not the final frontier for link prediction. More complex subgraph-GNN architectures \cite{zhang2018link, zhu2021neural, zhao2021stars} show superior link performance, and newer work show strong effectiveness of joint usage of simple heuristic link prediction metrics with GNNs \cite{mao2024demystifying, mamixture,zhao2022learning}. Moreover, co-training of large language models with graphs to improve node embedding quality is an emerging area~\cite{chenllaga,ren2024survey,zhao2024learning}.  
We see explorations in all these areas as avenues for the next-generation of GNNs at Snapchat.

\paragraph{Broadening Applications.} 
While our current applications mostly focus on link prediction for recommendation systems, significant potential exists in expanding into node-level classification and regression tasks, particularly for user modeling. Moreover, cross-domain graph modeling offers opportunities of knowledge transfer between high- and low-resource interaction domains, as well as social recommendation avenues which directly leverage social graph in the context of user-item recommendation tasks \cite{fan2019graph, zhu2021cross}.

\paragraph{Evolving Technical Choices.} We discussed key elements of GiGL design in \cref{sec:framework}. These choices enable flexibility in GNN training and inference which rely on in-memory subgraph sampling, graph databases, and ETL jobs. However, shifts in infrastructure, cloud offerings, and cost models can influence which approach is most effective for specific graph modeling tasks. We continue to explore these evolving opportunities to improve GiGL, and are eager to collaborate with the academic community to advance the field.

\vspace{-0.05in}
\section{Conclusion}
\label{sec:conclusion}

In this work, we introduced GiGL (Gigantic Graph Learning), an open-source library designed for scalable GNN training and inference on billion-scale graphs. Beyond the library itself, we provided insights into GiGL’s role in powering critical applications at Snapchat, contributing to over 35 product launches.
We also shared key lessons learned from developing and deploying GNNs at scale, covering topics from both infrastructure and modeling perspectives.
By open-sourcing GiGL, we aim to support the broader graph ML community while fostering collaboration around scalable graph learning and real-world GNN applications.

% \clearpage

\begin{acks}
We would like to thank Sam Young, Varun Nayini, Khai Tran, and Tyler Jiang (supporting OSS efforts), Neha Yadav, Crystal Wang, Ratna Kumar Kovvuri, Ovais Khan and Spoorth Ravi (infrastructure partners), Jiahui Shi, Yan Wu, Vivek Chaurasiya, Xiaohan Zhao, Nathaniel See, Chengjie Wu, and Lili Zhang  (product partners).
\end{acks}

\bibliographystyle{ACM-Reference-Format}
\bibliography{ref}

% \clearpage
% \onecolumn
\appendix
\section*{Appendices}

\begin{table*}[t]%[htp]
\centering
\caption{Component-level runtime comparison of GiGL on the same graph under different settings - under different resource and hyper-parameters. ``-'' indicates no change from baseline setup.}
\label{tab:resource}
\vspace{-0.1in}
\resizebox{0.99\textwidth}{!}{
\begin{tabular}{l|cccccc|c} 
\toprule
 & \multicolumn{6}{c|}{Runtime (mins)} & Offline \\
Setup & Total & Data Preprocessor & Subgraph Sampler & Split Generator & Trainer & Inferencer & MRR \\ 
\midrule
Baseline setup & 11h 24m & 2h 31m & 2h 29m & 50m & 4h 4m & 1h 30m & 0.820 \\
\cmidrule(lr){1-1} \cmidrule(lr){2-7} \cmidrule(lr){8-8}
Half training GPU & 14h 7m & - & - & - & 6h 47m & - & - \\
Half inference CPU & 11h 43m & - & - & - & - & 1h 49m & - \\
\cmidrule(lr){1-1} \cmidrule(lr){2-7} \cmidrule(lr){8-8}
Early stop patience * 2 & 19h 1m & - & - & - & 11h 41m & - & 0.858\\
\# of positive samples in training * 2 & 15h 58m & 2h 43m & 3h 58m & 1h 6m & 5h 43m & 1h 28m & 0.835 \\
\# of neighbors in subgraph sampling / 2 & 7h 29m & - & 1h 48m & 27m & 1h 43m & 1h & 0.786 \\
Remove edge features & 9h 27m & 2h 1m & 2h 25m & 46m & 2h 47m & 1h 28m & 0.806\\
\bottomrule
\end{tabular}
}
\vspace{-0.1in}
\end{table*}
\begin{table*}[t]
\centering
\caption{Statistics of datasets used in the offline experiments.}
\vspace{-0.1in}
\label{tab:datasets}
\begin{tabular}{l|cccc} 
\toprule
Dataset & \# Nodes & \# Edges & \# Node Features & \# Edge Features \\ 
\midrule
MAG240M & 244,134,778 & 1,683,771,646 & 768 & 0 \\
Internal-small & 3,572,811 & 125,897,892 & 249 & 19\\
Internal-less-feat & 899,382,946 & 16,840,453,931 & 155 & 10 \\
Internal-less-edge & 899,382,946 & 8,420,226,965 & 249 & 19 \\
Internal-full & 899,382,946 & 16,840,453,931 & 249 & 19\\
\bottomrule
\end{tabular}
\vspace{-0.1in}
\end{table*}

\begin{table}[htp]
\centering
\caption{Component-level runtime of GiGL on a internal heterogeneous graph, with the GraphDB backend.}
\label{tab:scale-het}
\vspace{-0.1in}
\begin{tabular}{l|c} 
\toprule
DataPreprocessor (mins) & 58m \\
Subgraph Sampler (mins) & 2h 52m \\
Split Generator (mins) & 9m \\
Trainer (mins) & 3h 31m \\
Inferencer (mins) & 59m \\
Total Runtime (mins) & 8h 29m\\
Offline MRR & 0.844 \\
\bottomrule
\end{tabular}
\vspace{-0.1in}
\end{table}

\section{Dataset Details}
\label{appx:dataset}

In order to showcase the impact of graph scale on GiGL runtime, we vary the internal graph in terms of number of nodes/edges and number of node/edge features. Detailed statistics of these datasets are included in \cref{tab:datasets}. The three smaller internal datasets (Internal-small, Internal-less-feat, and Internal-less-edge) are all sub-sampled from Internal-full.

For the experiments in \cref{tab:scale}, we used the same resource configuration to ensure a fair comparison. However, it's worth noting that the resource configuration can and should be further tuned for specific jobs, resulting with optimal cost and time efficiency.

\paragraph{Internal-full}. Snap internal large scale homogeneous graph dataset containing $\sim${900M} nodes, and $\sim${16.8B} edges. With 249 node features and 19 edge features.

\paragraph{Internal-small}. A small sample of the Snap internal dataset with $\sim${3.5M} nodes and $\sim$125M edges, with feature dimension unchanged.

\paragraph{Internal-less-edge}. We randomly remove $50\%$ of the existing edges in the Internal-full dataset, leaving $\sim$8.4B edges. \#nodes left and features unchanged.

\paragraph{Internal-less-feat}. We reduce the node feature dimension to 155, and edge feature dimension to 10. 

\paragraph{MAG240M}\footnote{\url{https://ogb.stanford.edu/docs/lsc/mag240m/}}. 
In order to evaluate the same link prediction task as we did for the internal datasets, the train/validation/test sets are generated by our Split Generator following a standard transductive link prediction split.
We leverage the original public academic graph dataset \cite{hu2021ogblsc, hu2020ogb} and cast it to a homogeneous graph for our experiments by: (i) using a 768 size zero vector as the feature vector for each author node. (ii) Combine the new hydrated author nodes and original paper nodes with 768 dim float features into a single node table. (iii) Add the node degree of each node as the 769th feature. (iv) Combine the paper-cites-paper and the author-writes-paper edge table to a single edge table.

\section{Additional Offline Experiments}
\label{appx:experiments}

In \cref{tab:scale-het}, we provide runtime and offline evaluations on an internal heterogeneous (bipartite) graph, with the GraphDB backend for Subgraph Sampler. This graph contains $\sim$114.6M nodes in one node type (482 features), $\sim$10.8M nodes in another node type (124), and $\sim$4.6B edges (5 edge features).

\subsection{Varying Level of Parallelism and Configurations}

We also conduct runtime experiments with different resource settings and hyperparameters. \cref{tab:resource} shows their component-level runtimes as well as offline MRR for link prediction. Following are the details of each setting:

\noindent $\bullet$ \textbf{Half training GPU}. Reduce the \# of training GPUs from 16 to 8. This affects training only, since inference is done with CPUs.

\noindent $\bullet$ \textbf{Half inference CPU}. Reduce the \# of inference CPUs by half. This only affects inferencer and not trainer or other components.

\noindent $\bullet$ \textbf{Early stop patience * 2}. Double the patience will result in better loss optimization and metric, but slower training time.

\noindent $\bullet$ \textbf{\# of positive samples * 2}. Double the positive labels sampled for each training sample. This increases Subgraph Sampler \& Trainer run time.

\noindent $\bullet$ \textbf{\# of neighbors in subgraph sampling / 2}. We reduce the fanout of the neighborhood subgraph at each hop by half. This improves multiple components' (mainly Subgraph Sampler) run time. 

\noindent $\bullet$ \textbf{Remove edge features}. We specify the config to remove the use of any edge features. This will in turn speed up all components. Removing edge features is more efficient than reducing node features, since edge are at the scale of billions in the internal dataset comparing millions for nodes. 

\section{Config Examples}
\label{appx:config}

\paragraph{Example Task Config} for GiGL pipeline on MAG240M dataset for the task of link prediction: \url{https://github.com/Snapchat/GiGL/blob/main/examples/MAG240M/task_config.yaml}. Detailed implementation of GiGL end-to-end pipeline on MAG240M dataset for link prediction is also provided as an example in the GiGL repository\url{https://github.com/Snapchat/GiGL/tree/main/examples/MAG240M}.

\paragraph{Example Resource Config} used with \verb+task_config.yaml+ specified above: \url{https://github.com/Snapchat/GiGL/blob/main/examples/MAG240M/resource_config.yaml}.

\end{document}